# Predicting Conditional Quantiles via Reduction to Classification

**John Langford**
Toyota Technological Institute
1427 E 60th Street
Chicago, IL 60637
jl@hunch.net

**Roberto Oliveira**
IBM T. J. Watson Research Center
P.O. Box 218
Yorktown Heights, NY 10598
rob.oliv@gmail.com

**Bianca Zadrozny**
Universidade Federal Fluminense
Rua Passo da Pátria 156, Bl. E, 3 andar
Niterói, RJ 24210-240, Brazil
bianca@ic.uff.br

## Abstract

We show how to reduce the process of predicting conditional quantiles (and the median in particular) to solving classification. The accompanying theoretical statement shows that the regret of the classifier bounds the regret of the quantile regression under a quantile loss. We also test this reduction empirically against existing quantile regression methods on large real-world datasets and discover that it provides state-of-the-art performance.

## 1 Introduction

Regression is the problem of estimating a mapping from some feature space $X$ to a real-valued output $Y$, given a finite sample of the form $(x, y)$ drawn from a distribution $D$ over $X \times Y$. Typically, the goal of regression is to minimize the squared-error loss over the distribution $D$, that is, $E_{(x,y)\sim D}(y - f(x))^2$. One standard justification for this form of regression is that the minimizer is the mean: $f^*(x) = E_{y \sim D|x}[y]$.

However, there are many important applications for which mean estimates are either irrelevant or insufficient, and *quantiles* (also known as general order statistics) are the main quantities of interest. For instance, consider trying to assess the risk of a business proposal. Estimates of the lower quantiles of the conditional return distribution would give a better indication of how worthwhile the proposal is than a simple estimate of the mean return (which could be too high because of very unlikely high profits).

The process of estimating the quantiles of a conditional distribution is known as *quantile regression*. More specifically, the goal of quantile regression is to obtain estimates on the $q$-quantiles of the conditional distribution $D|x$. Intuitively, $q$-quantiles for different $q$ describe different segments of the conditional distribution $D|x$ and thus offer more refined information about the data at hand. Other reasons for doing quantile regression as opposed or in addition to typical regression include:

1. Quantiles tend to behave well under noise. For instance, the median (i.e. the $1/2$-quantile) equals the mean under Gaussian noise, but the median is often inherently more robust to heavy tailed and non-Gaussian noise;

2. As mentioned above, many important practical problems are naturally expressed in terms of quantiles. For instance, *wallet estimation* – e.g. estimating the potential amount of money a customer can spend on computer hardware, as opposed to the expected amount – can be done by looking at the conditional upper quantiles of expenditures [10, 9]. Quantile regression also been applied to many other problems in Econometrics, Sociology and Ecology, among other fields [5, 6];

3. The actual distribution of conditional noise can be estimated as required for some applications [8] using quantile regression.

This paper shows that the quantile regression problem can be reduced to classification. The *Quanting* algorithm that we introduce takes as input an instance of quantile regression and outputs a family of classification problems such that solving the latter problems with small *average* error leads to a provably accurate estimate of the conditional quantile. Reducing quantile to classification automatically gives us access to a large array of quantile regression methods, since the reduction applies to any existing or future classification method.

We compare empirically the Quanting algorithm with other methods for quantile regression in the literature. Koenker [5] has developed a linear quantile regression method, while Takeuchi et al.[11] have recently devised a kernel-based quantile estimation method. Our approach, which is intrinsically non-linear and conceptually simpler, compares favorably with the existing alternatives in our experiments.

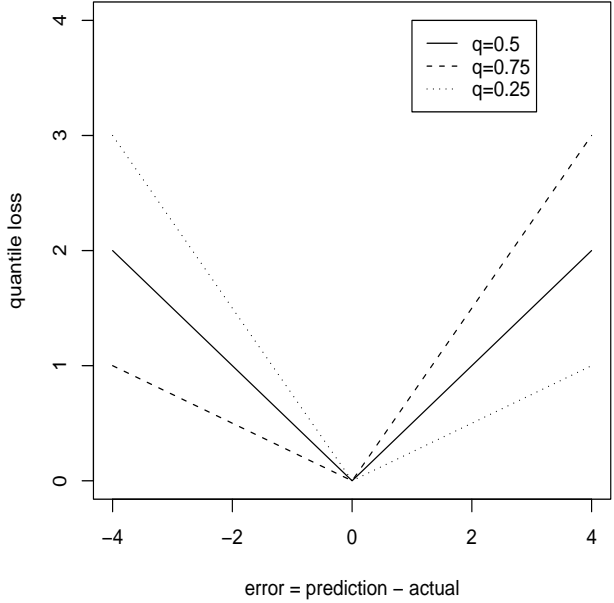

Figure 1: Loss functions which induce quantile regression.

## 2 Basic Details

The quantile regression problem is defined in a setting where we have a measure $D$ over a set of features $X$ and real-valued outputs $Y$.

**Definition 1** *(Conditional q-quantile)* Let $0 \leq q \leq 1$ $f = f(x)$ *is a* conditional $q$-quantile *(or* conditional $q$-order statistic*) for $D$ if for (D-almost every) $x \in X$ $D(y \leq f(x)|x) \geq q$ and $D(y \geq f(x)|x) \geq 1 - q$. The $1/2$-quantile is also known as the* median.

Note that the $q$-quantile may not be unique when the conditional distribution has regions with zero mass.

### 2.1 Optimization

It is well-known that the optimal estimator for the absolute-error loss is a median [6]. In other words, we have that for every regression problem $D$,

$$\arg \min_f E_{x,y \sim D}|y - f(x)| \text{ is a (conditional) median.}$$

This can be verified by considering two equal point masses at locations $y_1$ and $y_2$. The absolute value loss for a point $y \in [y_1, y_2]$ is $(y - y_1) + (y_2 - y) = (y_2 - y_1)$ which is constant independent of $y$; whereas any $y \notin [y_1, y_2]$ yields a larger value. Since we can take any distribution over $y$ and break into equal mass pairs with $y_2$ above and $y_1$ below the median, the absolute-error loss is minimized when $f(x)$ is a median.

The generalization of absolute-error loss for arbitrary order statistics is the quantile loss function, also known as "pin-ball loss" [11]. Pictorially, this is a tilted absolute loss as in figure 1. Mathematically, this is $E_{x,y \sim D} l_q(y, f(x))$, where

$$\begin{aligned} l_q(y, f(x)) &= q[y - f(x)]I(y \geq f(x)) \\ &\quad + (1-q)[f(x) - y]I(y < f(x)), \end{aligned} \quad (1)$$

and $I(\cdot) = 1$ if its argument is true and 0 otherwise.

---

**Algorithm 1** Quanting-train (importance-weighted classifier learning algorithm $A$, training set $S$, quantile $q$)

1. For $t$ in $[0, 1]$
   (a) $S_t = \{\}$;
   (b) For each $(x, y)$ in $S$:
       $S_t = S_t \cup \{(x, I(y \geq t), qI(y \geq t) + (1 - q)I(y < t))\}$
   (c) $c_t = A(S_t)$

2. Return the set of classifiers $\{c_t\}$

---

The correctness of this can be (again) seen by considering two points $y_1$ and $y_2$ with probability ratio $\frac{1-q}{q}$; or by noting that the result is implied by equation (3) in the proof of Lemma 1 (the integral is positive unless $Q(x) = q(x)$ or $D(y < t|x) = q$ for all $t$ between $Q(x)$ and $q(x)$).

## 3 An Algorithm

In this section and the next, we assume that we are given samples $(x, y)$ from a distribution $D$, where $0 \leq y \leq 1$ (if this is not the case, we can re-normalize the data). Given these samples, our proposed algorithm, which we call Quanting, estimates the $q$th quantile of the conditional distribution $D|x$ using any importance weighted classification algorithm $A$. In fact, using an extra reduction discussed in Corollary 1 below, one can also do Quanting via an unweighted binary classification algorithm, but we defer any further discussion of this to the next section.

The Quanting algorithm has two parts. Algorithm 1 receives a set of training examples $S$ of the form $(x, y)$ and a quantile $q$ as input and uses algorithm $A$ to compute a family of binary classifiers $c_t$. We assume that algorithm $A$ receives as input a set of training examples of the form $(x, y, w)$ where $w$ is a weight and attempts to minimize the weighted error. In the algorithm, positive examples receive weight $q$ while negative examples receive weight $(1 - q)$. Using the classifiers $c_t$, Algorithm 2 produces a prediction of the $q$-quantile for each $x$ in a test set $S'$, in precisely the same manner as the Probing algorithm for estimating conditional class probabilities using 0/1 classifiers [7].

The essential idea of Quanting is also similar to Probing. Each $c_t$ attempts to answer the question "is the $q$-quantile above or below $t$?" In the (idealized) scenario where $A$ is perfect, one would have $c_t(x) = 1$ if and only if $t \leq$

**Algorithm 2** Quanting-test (set of classifiers $\{c_t\}$, test set $S'$)

1. For each $x$ in $S'$:
   $Q(x) = E_{t \sim U(0,1)}[c_t(x)]$

---

$q(x)$ for a $q$-quantile $q(x)$, hence Algorithm 2 would output $\int_0^{q(x)} dt = q(x)$ exactly. Our analysis shows that if the error of $A$ is small on average over $t$, the quantile estimate is accurate.

We note in passing that in reality, one cannot find a different classifier for each $t \in [0, 1]$. Constructing classifiers $c_t$ for $t$ in a discrete mesh $\{0, 1/n, 2/n, \ldots, (n-1)/n, 1\}$ will add a $1/n$ term to the error bound.

## 4 Quanting Reductions Analysis

The Lemma we prove next relates the average regret of the classifiers $c_t$ (how well the classifiers do in comparison to how well they could do) to the regret of the quantile loss incurred by the Quanting algorithm. For each $x$, the output produced by the quanting algorithm is denoted by $Q(x)$, whereas $q(x)$ is a correct $q$-quantile. In this analysis, we use the standard one-classifier trick [7]: instead of learning different classifiers, we learn one classifier $c = \{c_t\}$ with an extra feature $t$ used to index classifier $c_t$.

**Lemma 1** *(Quanting Regret Transform) For all $D$, $c$,*

$$E_{x,y \sim D}[l_q(y, Q(x))] - E_{x,y \sim D}[l_q(y, q(x))]$$

$$\leq e(D, c) - \min_{c'} e(D, c')$$

*where $e(D, c)$ is the expected importance weighted binary loss of $c$ over $D$.*

*Proof:* For any function $f = f(x)$, $E_{x,y \sim D} l_q(y, f(x))$ is given by eqn. (1):

$$qE_{x,y \sim D}(y - f(x))I(y - f(x) > 0)$$
$$+ (1-q)E_{x,y \sim D}(f(x) - y)I(f(x) - y > 0).$$

It is known that $E[XI(X > 0)] = \int_0^{+\infty} \Pr(X \geq t)dt = \int_0^{+\infty} \Pr(X > t)dt$ for any random variable $X$, so we rewrite:

$$E_{x,y \sim D} l_q(y, f(x))$$
$$= qE_x \int_0^\infty D(y - f(x) \geq t_1|x) dt_1$$
$$+ (1-q)E_x \int_0^\infty D(f(x) - y > t_2|x) dt_2$$
$$= qE_x \int_{f(x)}^1 D(y \geq u|x) \, du$$
$$+ (1-q)E_x \int_0^{f(x)} D(y < u|x) \, du. \qquad (2)$$

Applying this formula to $f(x) = Q(x)$ and $f(x) = q(x)$ and taking the difference yields

$$E_{x,y \sim D}[l_q(y, Q(x)) - l_q(y, q(x))]$$
$$= E_x \int_{Q(x)}^{q(x)} \begin{bmatrix} qD(y \geq u|x) \\ -(1-q)D(y < u|x) \end{bmatrix} du$$
$$= E_x \int_{Q(x)}^{q(x)} \begin{bmatrix} q - qD(y < u|x) \\ -(1-q)D(y < u|x) \end{bmatrix} du$$
$$= E_x \int_{Q(x)}^{q(x)} [q - D(y < u|x)] \, du. \qquad (3)$$

We will show that $e(D, c) - \min_{c'} e(D, c')$ is at least this last expression. The expected importance-weighted error incurred by the classifiers $\{c_t\}$ is

$$e(D, c) = E_{x,y \sim D} \int_0^1 \begin{bmatrix} qI(y \geq t)(1 - c_t(x)) \\ +(1-q)I(y < t)c_t(x) \end{bmatrix} dt$$
$$= E_x \int_0^1 \begin{bmatrix} qD(y \geq t|x) \\ +(D(y < t|x) - q)c_t(x) \end{bmatrix} dt$$
$$= qE_x[y] + E_x \int_0^1 [D(y < t|x) - q]c_t(x) \, dt \quad (4)$$
$$\geq qE_x[y] + E_x \int_0^{Q(x)} [D(y < t|x) - q] dt. \quad (5)$$

Here only the last line is non-trivial, and it follows from the fact that $D(y < t|x) - q$ is increasing in $t$. Thus the smallest possible value for the integral in (4) is achieved by placing as much "weight" $c_t(x)$ as possible on the smallest $t$ while respecting the constraints $\int_0^1 c_t(x) \, dt = Q(x)$ and $0 \leq c_t(x) \leq 1$. This corresponds precisely to setting $c_t(x) = I(t \leq Q(x))$, from which (5) follows.

On can show that the inequality (5) is in fact an equality when instead of $\{c_t\}$ we use the (optimal) classifiers

$$\{c_t^*(x) = I(D(y \leq t|x) \leq q)\}$$

and substitute $q(x)$ for $Q(x)$. Therefore,

$$e(D, c) - e(D, c^*)$$
$$\geq E_x \int_{q(x)}^{Q(x)} [D(y < t|x) - q] dt$$
$$= E_{x,y \sim D}[l_q(y, Q(x)) - l_q(y, q(x))],$$

using (3). This finishes the proof. ∎

We now show how to reduce $q$-quantile estimation to unweighted binary classification using the results of previous work [2]. We apply *rejection sampling*: we feed the unweighted classifier samples of the form $((x, t), I(y \geq t))$, each of the samples being independently discarded with probability $1 - w(I(y \geq t))$, where $w(b) = qb + (1-q)(1-b)$ is the example's weight. Notice that by [2, Theorem 2.3], sample complexity is not significantly affected by this.

**Corollary 1** *(Quanting to Binary Regret) For $D$ as above and unweighted binary classifier $c = \{c_t\}$, let $\tilde{D}$ be the distribution produced by rejection sampling. Then*

$$E_{x,y \sim D}[l_q(y, Q(x))] - E_{x,y \sim D}[l_q(y, q(x))]$$

$$\leq e(\tilde{D}, c) - \min_{c'} e(\tilde{D}, c').$$

*Proof:* Let $\hat{c} = \{\hat{c}_t\}_t$ be the importance-weighted classifiers induced by the rejection sampling procedure. A folk theorem [2] implies that

$$e(D, \hat{c}) - \min_{\hat{c}'} e(D, \hat{c}') = e(\tilde{D}, c) - \min_{c'} e(\tilde{D}, c')$$

and the result follows from Lemma 1. ∎

## 5 Related Work

A standard technique for quantile regression that has been developed and extensively applied in the Econometrics community [6] is linear quantile regression. In linear quantile regression, we assume that the conditional quantile function is a linear function of the features of the form $\beta x$ and we estimate the parameters $\hat{\beta}$ that minimize the quantile loss function (Equation 1). It can be shown that this minimization is a linear programming problem and that it can be efficiently solved using interior point techniques [5]. Implementations of linear quantile regression are available in standard statistical analysis packages such as R and SAS.

The obvious limitation of linear quantile regression is that the assumption of a linear relationship between the explanatory variables and the conditional quantile function may not be true. Recently, Takeuchi et al. have recently proposed a technique for nonparametric quantile estimation [11] that applies the two standard features of kernel methods to conditional quantile estimation: regularization and the kernel trick. They show that a regularized version of the quantile loss function can be directly minimized using standard quadratic programming techniques. By choosing an appropriate kernel, such as a radial basis function kernel, one can obtain nonlinear conditional quantile estimates. They compare their method experimentally to linear quantile regression and to a nonlinear spline approach suggested by Koenker [5] on many small datasets for different quantiles and find that it performs the best in most cases.

## 6 Experiments

Here we compare experimentally the Quanting algorithm to the two existing methods for quantile regression described in section 5: linear quantile regression and kernel quantile regression.

We compare the methods on three different performance metrics:

1. The quantile loss (Equation 1).

2. The percentage of examples for which the prediction $f(x)$ exceeds the actual value $y$, which should be close to the quantile $q$ for which we are optimizing.

3. The running time (Pentium 1.86GHz, 1.00GB RAM).

As base classifier learners for Quanting, we use two algorithms available in the WEKA machine learning software [13]: the J48 decision tree learner and logistic regression. For both methods, we use the default parameters provided by WEKA. We use rejection sampling to perform importance-weighted classification with standard unweighted classifiers. We use an adaptive discretization scheme to choose the thresholds $t$ in algorithm 1, the same scheme that we use in the Probing reduction [7]. We fix the number of classifiers at 100 for all the datasets.

For the kernel quantile regression, we have followed the same experimental setup as described by Takeuchi et al. [11]. We use a radial basis function kernel and choose its radius and the regularization parameter using cross-validation on the training data. We also scale the features and the label to have zero mean and a standard deviation of 1, as required for kernel methods. When predicting, we convert the label back to the original scale to compute the quantile loss.

As our benchmarks, we use four large, publicly available datasets, from real-world domains where quantile regression is clearly applicable:

1. Adult: available from the UCI Machine Learning Repository [4] as a classification dataset. The data was originally extracted from the Census Bureau Database and describes individual demographic characteristics of such as age, education, sex and occupation. For the original dataset, the objective is to predict a label that indicates whether or not the individual's income is above $50K. We have retrieved the original numerical income values from the Census Bureau Database and used the income as the dependent variable in the quantile regression. Our objective is to predict the quantiles of the conditional income distribution, which is useful if we want to determine what is a low or a high income for a given individual.

2. KDD-Cup 1998: available from the UCI KDD Archive [3]. This dataset consists of records of individuals who have made a donation in the past to a particular charity. Each example consists of attributes describing each individual's donation history over a series of donation campaigns, as well as demographic information, such as income and age. The dependent variable is the individual's donation amount in the most recent donation campaign. The original dataset contains 95412 training records and 96367 test records, but only 5% of the individuals donated in the current campaign. Our objective is to predict the quantiles of the conditional donation amount for individuals who donate. For this reason, we only use 4843 donor examples in the training set and the 4876

| Dataset | Features | Training | Test |
|---|---|---|---|
| Adult | 14 | 32560 | 16280 |
| KDD-Cup 1998 | 10 | 4840 | 4870 |
| California Housing | 8 | 13760 | 6880 |
| Boston Housing | 14 | 450 | 56 |

Table 1: Number of features and examples (training and test) of each of the datasets.

   donor examples in the test set. Predicting quantiles of the conditional donation distribution is important for "anchoring", i.e., deciding how much to suggest as possible donation values when soliciting donations. Anchoring is a well-established concept in marketing, see, for example, [12].

3. California Housing: available from the StatLib repository [1]. It contains data on California housing characteristics aggregated at the block level (a sample block group on average includes 1425.5 individuals living in a geographically compact area). The independent variables are median income, housing median age, total rooms, total bedrooms, population, households, latitude, and longitude. The dependent variable is the median house value. Our objective is to predict the quantiles of the conditional house value distribution. This information is very valuable for house sellers and buyers, since it indicates what would be a 'lower bound' and an 'upper bound' on the house value, given its characteristics.

4. Boston Housing: available from the StatLib repository [1]. It contains data on Boston housing characteristics and values. The prediction task is analogous to the one described for the California Housing dataset.

Table 1 shows the number of features and the number of examples for each dataset. We use the standard train/test splits for training and testing. Because the kernel quantile regression method has very high memory and computational time requirements, we could not run it using all the examples in the training set for the Adult, KDD-Cup 1998 and California Housing. We have run it for the maximum number of examples possible, which in this case was 3000 for the three datasets (after trying with 1000, 2000, 3000, 4000, etc.). The 3000 examples were chosen at random from the training data.

We have run the methods for 3 different quantile values for each dataset: 0.1, 0.5 and 0.9. The results are shown in tables 2, 3, 4 and 5.

In terms of running time, it is clear that the linear method is the most efficient and that the kernel method is inefficient. Even with the number of examples limited at 3000, the kernel method takes more than one hour to run on the larger datasets. Quanting is relatively efficient: with our choice of classifier learners it does not take more than 8 minutes to run with 100 classifiers even on the larger datasets. In terms of the quantile loss, Quanting-J48 is clearly the best performer for the Adult and California Housing datasets. The kernel method and Quanting-LogReg are performing about the same for these two datasets, while the linear method is inferior. For the KDD-Cup 1998 dataset, the linear method is the best for q=0.5 and q=0.9, while Quanting-J48 is the best for q=0.1. This can be explained by the fact that there is a strong linear correlation between the label and one of the features in this dataset, which is well-captured by the linear quantile regression but not so easily captured by Quanting and by the kernel method. Finally, for the Boston Housing dataset, the best method depends on the particular value of $q$ but the linear method is performing consistently worse than the others.

In terms of the percentage of examples for which the prediction exceeds the actual value, all the methods come close to desired value (the same as $q$) in most of the cases. But we can observe that the linear method is consistently close, while the kernel method shows the largest deviations.

To give an idea of how the Quanting algorithm progresses as we add more classifiers, in figure 2 we plot the quantile loss as a function of the number of classifiers for the Adult and Boston Housing datasets ($q = 0.9$). For comparison, we also plot the values of the quantile loss for the linear and kernel methods as horizontal lines in the picture. It is clear that Quanting converges very fast. In both cases, the convergence occurs with about 50 classifiers.

## 7 Conclusion

In this paper, we present a reduction from quantile regression to classification. Theoretically, we are now able to quantify the regret of quantile regression under a quantile loss in terms of the error rate of a base classifier. In practice, this means that we can apply classifier learning methods to solve quantile regression problems, which appear very often in real-world applications. Our experiments show that the Quanting reduction is efficient in terms of computational time and performs well compared to existing quantile regression methods.

### Acknowledgements

We thank Saharon Rosset and Claudia Perlich for useful discussions on the topic of this paper.

## References

[1] StatLib: Data, Software and News from the Statistics Community. Department of Statistics, Carnegie Melon University, 2006. http://lib.stat.cmu.edu/.

[2] Naoki Abe, Bianca Zadrozny, and John Langford. Cost sensitive learning by cost-proportionate example weighting. In

| $q = 0.1$ | | | |
|---|---|---|---|
| Method | Quantile Loss | % Below | Running Time |
| linear | 2527.94 | 0.1030035 | **7.48s** |
| kernel | 2301.98 | 0.1102512 | 91m40.280s |
| quanting-LogReg | 2307.17 | 0.0969227 | 6m28.440s |
| quanting-J48 | **1988.79** | **0.1010382** | 10m0.839s |

| $q = 0.5$ | | | |
|---|---|---|---|
| Method | Quantile Loss | % Below | Running Time |
| linear | 6903.85 | 0.4919231 | **14.66s** |
| kernel | 6038.98 | **0.4957926** | 42m32.093s |
| quanting-LogReg | 6059.53 | 0.4730667 | 7m6.513s |
| quanting-J48 | **5206.45** | 0.4956083 | 12m17.903s |

| $q = 0.9$ | | | |
|---|---|---|---|
| Method | Quantile Loss | % Below | Running Time |
| linear | 3332.05 | 0.89601 | **7.04s** |
| kernel | 3198.38 | **0.89859** | 116m30.499s |
| quanting-LogReg | 3029.29 | 0.86893 | 5m54.691s |
| quanting-J48 | **2698.93** | 0.87992 | 10m48.305s |

Table 2: Results on the Adult Dataset for $q = 0.1$, $q = 0.5$ and $q = 0.9$. The bold-face indicates the best result for each of the metrics.

| $q = 0.1$ | | | |
|---|---|---|---|
| Method | Quantile Loss | % Below | Running Time |
| linear | 0.930922 | **0.10158** | **2.37s** |
| kernel | 0.905908 | 0.058896 | 56m18.874s |
| quanting-LogReg | 0.962085 | 0.089062 | 6m42.427s |
| quanting-J48 | **0.895997** | 0.119844 | 1m5.388s |

| $q = 0.5$ | | | |
|---|---|---|---|
| Method | Quantile Loss | % Below | Running Time |
| linear | **1.926596** | 0.500513 | **1.25s** |
| kernel | 2.054151 | 0.413913 | 50m20.092s |
| quanting-LogReg | 1.996216 | 0.484301 | 1m58.684s |
| quanting-J48 | 1.929433 | 0.519803 | 1m9.559s |

| $q = 0.9$ | | | |
|---|---|---|---|
| Method | Quantile Loss | % Below | Running Time |
| linear | **1.222342** | 0.905397 | **12.66s** |
| kernel | 1.284019 | 0.877283 | 95m15.952s |
| quanting-LogReg | 1.355031 | 0.894316 | 1m35.620s |
| quanting-J48 | 1.332344 | **0.902319** | 1m1.079s |

Table 3: Results on the KDD-Cup 1998 Dataset for $q = 0.1$, $q = 0.5$ and $q = 0.9$. The bold-face indicates the best result for each of the metrics.

| $q=0.1$ | | | |
|---|---|---|---|
| Method | Quantile Loss | % Below | Running Time |
| linear | 8916.72 | **0.100581** | **2.28s** |
| kernel | 10063.12 | 0.169331 | 39m35.921s |
| quanting-LogReg | 8379.75 | 0.096512 | 2m25.617s |
| quanting-J48 | **6951.29** | 0.099128 | 2m25.168s |

| $q=0.5$ | | | |
|---|---|---|---|
| Method | Quantile Loss | % Below | Running Time |
| linear | 24627.62 | **0.496802** | **3.64s** |
| kernel | 23125.53 | 0.511192 | 36m54.890s |
| quanting-LogReg | 21617.22 | 0.515262 | 2m25.926s |
| quanting-J48 | **16195.01** | 0.522674 | 2m44.170s |

| $q=0.9$ | | | |
|---|---|---|---|
| Method | Quantile Loss | % Below | Running Time |
| linear | 13901.9 | **0.901017** | **2.86s** |
| kernel | 12905.64 | 0.851017 | 39m41.671s |
| quanting-LogReg | 11932.31 | 0.883430 | 2m7.973s |
| quanting-J48 | **9740.11** | 0.897819 | 2m14.491s |

Table 4: Results on the California Housing Dataset for $q=0.1$, $q=0.5$ and $q=0.9$. The bold-face indicates the best result for each of the metrics.

| $q=0.1$ | | | |
|---|---|---|---|
| Method | Quantile Loss | % Below | Running Time |
| linear | 0.659609 | 0.071429 | **0.09s** |
| kernel | **0.646979** | 0.071429 | 13m55.124s |
| quanting-LogReg | 0.662209 | 0.142857 | 0m33.966s |
| quanting-J48 | 0.779306 | **0.107143** | 0m24.856s |

| $q=0.5$ | | | |
|---|---|---|---|
| Method | Quantile Loss | % Below | Running Time |
| linear | 1.369074 | 0.410714 | **0.16s** |
| kernel | 1.327741 | 0.285714 | 11m50.452s |
| quanting-LogReg | **1.028511** | **0.535714** | 3m26.441s |
| quanting-J48 | 1.236357 | 0.392857 | 26.405s |

| $q=0.9$ | | | |
|---|---|---|---|
| Method | Quantile Loss | % Below | Running Time |
| linear | 1.137319 | **0.892857** | **0.13s** |
| kernel | 0.632955 | 0.946429 | 14m26.624s |
| quanting-LogReg | 0.693948 | 0.928571 | 34.161s |
| quanting-J48 | **0.580906** | 0.910714 | 59.594s |

Table 5: Results on the Boston Housing Dataset for $q=0.1$, $q=0.5$ and $q=0.9$. The bold-face indicates the best result for each of the metrics.

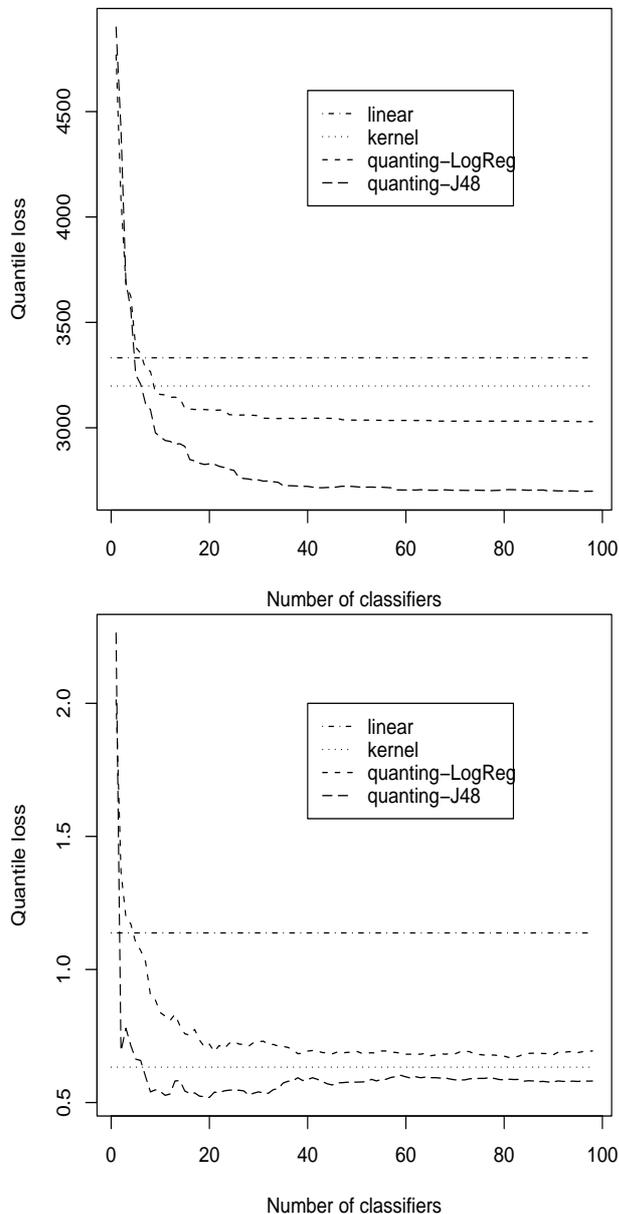

Figure 2: Progress of quanting on the Adult dataset (above) and Boston dataset (below) as the number of classifiers grows. In both cases $q = 0.9$.

*Proceedings of the 2003 IEEE International Conference on Data Mining (ICDM'03)*, pages 435–442, 2003.

[3] S. D. Bay. UCI KDD archive. Department of Information and Computer Sciences, University of California, Irvine, 2006. http://kdd.ics.uci.edu/.

[4] C. L. Blake and C. J. Merz. UCI repository of machine learning databases. Department of Information and Computer Sciences, University of California, Irvine, 2006. http://www.ics.uci.edu/∼mlearn/MLRepository.html.

[5] Roger Koenker. *Quantile Regression*. Econometric Society Monograph Series. Cambridge University Press, 2005.

[6] Roger Koenker and Kevin Hallock. Quantile regression. *Journal of Economic Perspectives*, 15:143–156, 2001.

[7] John Langford and Bianca Zadrozny. Estimating class membership probabilities using classifier learners. In *Proceedings of the Tenth International Workshop on Artificial Intelligence and Statistics*, pages 198–205. Society for Artificial Intelligence and Statistics, 2005.

[8] Joaquin Quinonero-Candela, Carl Edward Rasmussen, Fabian Sinz, Olivier Bousquet, and Bernhard Schlkopf. Evaluating predictive uncertainty challenge. In *Evaluating Predictive Uncertainty, Visual Object Categorization and Textual Entailment*, volume 3944 of *Lecture Notes in Computer Science*, pages 1–27. Springer Verlag, 2006.

[9] Saharon Rosset, Claudia Perlich, and Bianca Zadrozny. High-quantile modeling for marketing, sales and outlier detection applications. Submitted to KDD-2006, 2006.

[10] Saharon Rosset, Claudia Perlich, Bianca Zadrozny, Srujana Merugu, Sholom Weiss, and Rick Lawrence. Wallet estimation models. In *Proceedings of the International Workshop on Customer Relationship Management: Data Mining Meets Marketing (CRM Workshop)*, 2005.

[11] Ichiro Takeuchi, Quoc V. Le, Tim Sears, and Alex Smola. Nonparametric quantile regression. NICTA Technical Report, 2005.

[12] Brian Wansink, Robert J. Kent, and Stephen J. Hoch. An anchoring and adjustment model of purchase quantity decisions. *Journal of Marketing Research*, 35(1):71–81, February 1998.

[13] Ian H. Witten and Eibe Frank. *Data Mining: Practical machine learning tools and techniques*. Morgan Kaufmann, San Francisco, 2005.